%% file: main.tex
\documentclass[letterpaper, 10 pt, conference]{ieeeconf}  
\usepackage[noadjust]{cite}

\IEEEoverridecommandlockouts                              

\usepackage[utf8]{inputenc} 
\usepackage[T1]{fontenc}    
\usepackage[pagebackref=true,breaklinks=true,letterpaper=true,colorlinks,bookmarks=false,linkcolor=red,urlcolor=blue]{hyperref}
\usepackage{url}            
\usepackage{booktabs}       
\usepackage{amsfonts}       
\usepackage{nicefrac}       
\usepackage{microtype}      
\usepackage{epsfig}
\usepackage{times}
\usepackage{amsmath}
\usepackage{amssymb}
\usepackage{makecell}
\usepackage{graphicx}
\usepackage{xspace}
\usepackage{paralist}
\usepackage{comment}
\usepackage{mathtools}
\usepackage{xcolor}
\usepackage{ragged2e}
\usepackage{subcaption}
\usepackage{commath}
\usepackage[capitalise,noabbrev,nameinlink]{cleveref}
\usepackage{gensymb}
\usepackage{booktabs}
\usepackage{xparse}
\usepackage{mathtools}
\usepackage{etoolbox}
\usepackage{setspace}
\usepackage{float}
\usepackage{algorithm}
\usepackage[noend]{algpseudocode}
\usepackage{bm}
\usepackage{mwe}

\graphicspath{{imgs/}}

\DeclarePairedDelimiterX{\SquareBrackets}[1]{[}{]}{#1}
\DeclarePairedDelimiterX{\RoundBrackets}[1]{(}{)}{#1}
\DeclarePairedDelimiterX{\DivergenceBrackets}[2]{[}{]}{#1\;\delimsize\|\;#2}

\NewDocumentCommand{\pr}{ O{p} r() }{
  \def\prArg{#2}\patchcmd{\prArg}{|}{\mid}{}{}#1\RoundBrackets{\prArg}}
\NewDocumentCommand{\p}{ r() }{\pr[p](#1)}
\NewDocumentCommand{\q}{ r() }{\pr[q](#1)}
\NewDocumentCommand{\Normal}{ r() }{\pr[\operatorname{Normal}](#1)}
\NewDocumentCommand{\Cat}{ r() }{\pr[\operatorname{Cat}](#1)}
\NewDocumentCommand{\Bin}{ r() }{\pr[\operatorname{Bin}](#1)}
\NewDocumentCommand{\Beta}{ r() }{\pr[\operatorname{Beta}](#1)}
\NewDocumentCommand{\Bernoulli}{ r() }{\pr[\operatorname{Bernoulli}](#1)}
\NewDocumentCommand{\Dir}{ r() }{\pr[\operatorname{Dir}](#1)}

\def\eqref#1{equation~\ref{#1}}

\def\1{\bm{1}}

\DeclareMathAlphabet{\mathsfit}{\encodingdefault}{\sfdefault}{m}{sl}
\SetMathAlphabet{\mathsfit}{bold}{\encodingdefault}{\sfdefault}{bx}{n}

\title{\LARGE \bf Learning to Augment Synthetic Images\\ for Sim2Real Policy Transfer}

\author{Alexander Pashevich$^{*,1}$, Robin Strudel$^{*,2}$, Igor Kalevatykh$^{2}$, Ivan Laptev$^{2}$, Cordelia Schmid$^{1}$}

\begin{document}
\maketitle

\footnotetext[1]{University Grenoble Alpes, Inria, CNRS, Grenoble INP, LJK, 38000 Grenoble, France.}
\footnotetext[2]{Inria, \'{E}cole normale sup\'{e}rieure, CNRS, PSL Research University, 75005 Paris, France.}
\renewcommand*{\thefootnote}{\fnsymbol{footnote}}
\footnotetext[1]{Equal contribution.}
\vspace*{-0.6cm}

\thispagestyle{empty}
\pagestyle{empty}

\input{sections/abstract.tex}
\input{sections/intro.tex}
\input{sections/related.tex}
\input{sections/approach.tex}
\input{sections/results.tex}
\input{sections/conclusion.tex}


\bibliographystyle{plain}

\end{document}

%% file: sections/abstract.tex
\begin{abstract}

Vision and learning have made significant progress that could improve robotics policies for complex tasks and environments.
Learning deep neural networks for image understanding, however, requires large amounts of domain-specific visual data.
While collecting such data from real robots is possible, such an approach limits the scalability as learning  policies typically requires thousands of trials.

In this work we attempt to learn manipulation policies in simulated environments.
Simulators enable scalability and provide access to the underlying world state during training.
Policies learned in simulators, however, do not transfer well to real scenes given the domain gap between real and synthetic data.
We follow recent work on domain randomization and augment synthetic images with sequences of random transformations.
Our main contribution is to {\em optimize} the augmentation strategy for sim2real transfer and to enable domain-independent policy learning.
We design an efficient search for depth image augmentations using object localization as a proxy task.
Given the resulting sequence of random transformations, we use it to augment synthetic depth images during policy learning.
Our augmentation strategy is policy-independent and enables policy learning with no real images.
We demonstrate our approach to significantly improve accuracy on three manipulation tasks evaluated on a real robot.

\end{abstract}

%% file: sections/intro.tex
\section{INTRODUCTION}

Learning visuomotor control policies holds much potential for addressing complex robotics tasks in unstructured and dynamic environments.
In particular, recent progress in computer vision and deep learning motivates new methods combining learning-based vision and control.
Successful methods in computer vision share similar neural network architectures, but learn {\em task-specific} visual representations, e.g.~for object detection, image segmentation or human pose estimation.
Guided by this experience, one can assume that successful integration of vision and control will require learning of policy-specific visual representations for particular classes of robotics tasks.

Learning visual representations requires large amounts of training data.
Previous work has addressed policy learning for simple tasks using real robots e.g., in~\cite{Levine2016LearningCollection,Pinto2016,Zhang2018DeepIL}.
Given the large number of required attempts (e.g.~800,000 grasps collected in~\cite{Levine2016LearningCollection}), learning with real robots might be difficult to scale to more complex tasks and environments.
On the other hand, physics simulators and graphics engines provide an attractive alternative due to the simple parallelization and scaling to multiple environments as well as due to access to the underlying world state during training. 

Learning in simulators, however, comes at the cost of the reality gap.
The difficulty of synthesizing realistic interactions and visual appearance typically induces biases and results in low performance of learned policies in real scenes.
Among several approaches to address this problem, recent work proposes domain randomization~\cite{Tobin2017DomainWorld,Tobin2018DomainRA} by augmenting synthetic data with random transformations such as random object shapes and textures.
While demonstrating encouraging results for transferring simlator-trained policies to real environments (``sim2real'' transfer), the optimality and generality of proposed transformations remains open.

\input{figures/teaser.tex}

\input{figures/overview.tex}

In this work we follow the domain randomization approach and propose to learn transformations optimizing sim2real transfer.
Given two domains, our method finds policy-independent sequences of random transformations that can be used to learn multiple tasks.
While domain randomization can be applied to different stages of a simulator, our goal here is the efficient learning of visual representations for manipulation tasks.
We therefore learn parameters of random transformations to bridge the domain gap between synthetic and real images. 
We here investigate the transfer of policies learned for depth images. However, our method should generalize to RGB inputs. 
Examples of our synthetic and real depth images used to train and test the ``Cup placing'' policy are illustrated in Fig.~\ref{fig:teaser}.

In more details, our method uses a proxy task of predicting object locations in a robot scene.
We synthesize a large number of depth images with objects and train a CNN regressor estimating object locations after applying a given sequence of random transformations to synthetic images.
We then score the parameters of current transformations by evaluating CNN location prediction on pre-recorded real images.
Inspired by the recent success of AlphaGo~\cite{Silver2017MasteringTG} we adopt Monte-Carlo Tree Search~(MCTS)~\cite{Coulom2006EfficientSA}
as an efficient search strategy for transformation parameters.

We evaluate the optimized sequences of random transformations by applying them to simulator-based policy learning.
We demonstrate the successful transfer of such policies to real robot scenes while using no real images for policy training.
Our method is shown to generalize to multiple policies.
The overview of our approach is illustrated in Fig.~\ref{fig:overview}.
The code of the project is publicly available at  the project website\footnote{\url{http://pascal.inrialpes.fr/data2/sim2real}}.

%% file: figures/teaser.tex
\begin{figure}[t]
\centering
 \includegraphics[width=\linewidth]{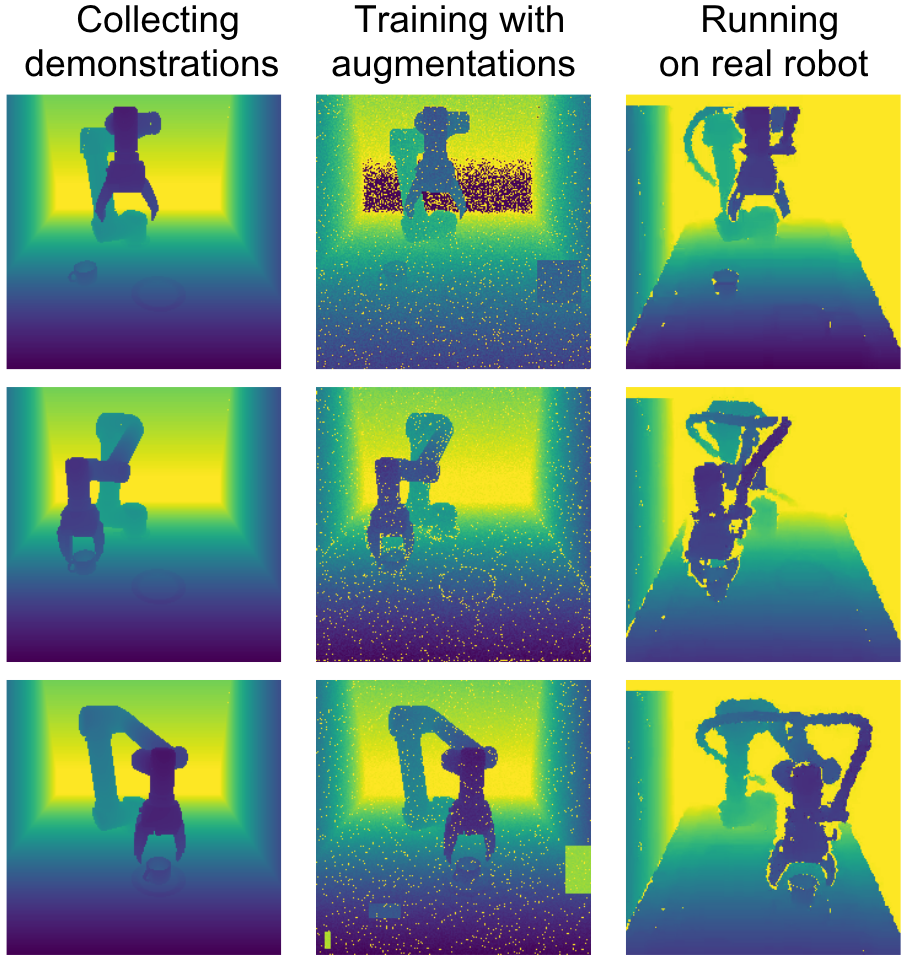}
\caption{Example depth images for the task ``Cup placing''. Synthetic depth maps (first column) are augmented with random transformations during policy training (second column). The resulting policy is applied to depth maps from the real robot scene (third column).}
\label{fig:teaser}
\end{figure}

%% file: figures/overview.tex
\begin{figure*}[t]
\centering
 \includegraphics[width=\linewidth]{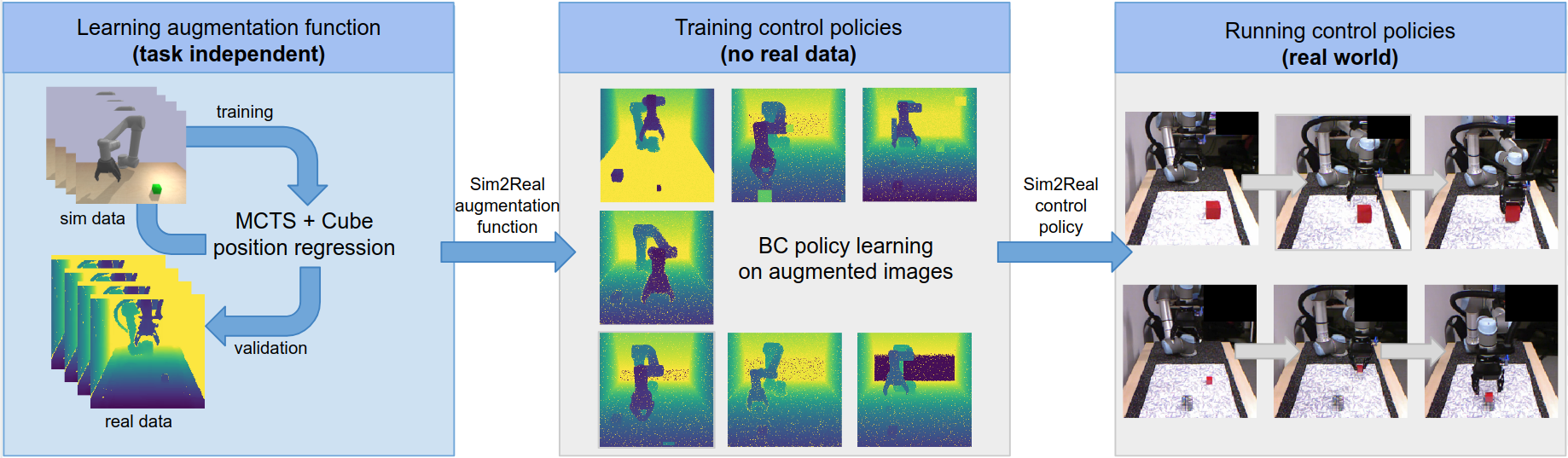}
\caption{Overview of the method. Our contribution is the policy-independent learning of depth image augmentations (left). The resulting sequence of augmentations is applied to synthetic depth images while learning manipulation policies in a simulator (middle). The learned policies are directly applied to real robot scenes without finetuning on real images.}
\label{fig:overview}
\end{figure*}

%% file: sections/related.tex
\section{RELATED WORK}

Robotics tasks have been addressed by various learning methods including imitation learning~\cite{2017arXiv170307326D} and reinforcement learning~\cite{SACX}.
Despite mastering complex simulated tasks such as Go~\cite{Silver2017MasteringTG} and Dota~\cite{OpenAI_dota}, the addressed robotics tasks remain rather simple~\cite{Zhang2018DeepIL, SACX, Gu2016}. This  difference is mainly caused by the real world training cost. Learning a manipulation task typically requires a large amount of data~\cite{Levine2016LearningCollection, Pinto2016}. Thus, robot interaction time becomes much longer than in simulation~\cite{Gu2016} and expert guidance is non-trivial~\cite{Zhang2018DeepIL}.
    
Learning control policies in simulation and transferring them to the real world is a potential solution to address these difficulties.  However, the visual input in simulation is significantly different from the real world and therefore requires adaptation~\cite{Sadeghi2017CAD2RLRS}.
Recent attempts to bridge the gap between simulated and real images can be generally divided into two categories: domain adaptation~\cite{Bousmalis2018UsingSA} and domain randomization~\cite{Tobin2017DomainWorld}. Domain adaptation methods either map both image spaces into a common one~\cite{James2018SimtoRealVS, Mueller2018DrivingPT} or map one into the other~\cite{Lee2018SPIGANPA}.
Domain randomization methods add noise to the synthetic images~\cite{Pinto2018AsymmetricAC, Sadeghi2017Sim2RealVI}, thus making the control policy robust to different textures and lighting.
The second line of work is attractive due to its effectiveness and simplicity. Yet, it was so far only shown to work with RGB images. 
While depth images are well suited for many robotics tasks~\cite{Litvak2018LearningAH}, it is not obvious what type of randomization should be used in the case of depth data. Here, we explore a learning based approach to select appropriate transformations and show that this allows us to close the gap between simulated and real visual data. 

Domain randomization is also referred to as data augmentation in the context of image classification and object detection. Data augmentation is known to be an important tool for training deep neural networks and in most cases it is based on a manually designed set of simple transformations such as mirroring, cropping and color perturbations.
In general, designing an effective data augmentation pipeline requires domain-specific knowledge~\cite{dvornik:hal-01844474}. Depth images might be augmented by adding random noise~\cite{ankur2016}, noise patterns typical for real sensors~\cite{Eitel2015MultimodalDL} or by compensating missing information~\cite{yang2012}.
Learning to augment is a scalable and promising direction that has been explored for visual recognition in~\cite{mattis2014}. Recent attempts to automatically find the best augmentation functions propose to use Reinforcement Learning and require several hundreds of GPUs~\cite{autoaugment}. 
Given the prohibitive cost of executing thousands of policies in a real-robot training loop, we propose to optimize sequences of augmentations within a proxy task by predicting object locations in pre-recorded real images using the Monte Carlo Tree Search~\cite{Coulom2006EfficientSA}.
A related idea of learning rendering parameters of a simulator has been recently proposed for a different task of semantic image segmentation  in~\cite{learningsim2019}.

%% file: sections/approach.tex
\section{APPROACH}
\label{sec:approach}

We describe the proposed method for learning depth image augmentations in Sections~\ref{sec:sim2real_eval} and \ref{sec:augment_search_space}.
Our method builds on Behaviour Cloning (BC) policy learning~\cite{Pomerleau1989,Ross2014} which we overview in Section~\ref{sec:approach_bc}.

\input{figures/transforms.tex}

\subsection{Behaviour cloning in simulation}
\label{sec:approach_bc}
Given a dataset $\mathcal{D}^{\textrm{expert}} = \{(o_t, a_t)\}$ of observation-action pairs along with the expert trajectories in \textit{simulation}, we learn a function approximating the conditional distribution of the expert policy $\pi_{\textrm{expert}}(a_t|o_t)$ controlling a robotic arm. Here, the observation is a sequence of the three last depth frames, $o_t = (I_{t-2}, I_{t-1}, I_t)  \in \mathcal{O} = \mathbb{R}^{H \times W \times 3}$. The action $a_t \in \mathcal{A} = \mathbb{R}^7$ is the robot command controlling the end-effector state. The action $a_t = (\mathbf{v}_t, \bm{\omega}_t, g_t)$ is composed of the end-effector linear velocity $\mathbf{v}_t \in \mathbb{R}^3$, end-effector angular velocity $\bm{\omega}_t \in \mathbb{R}^3$ and the gripper openness state $g_t \in \{0, 1\}$. 
We learn the deterministic control policy $\pi: \mathcal{O} \to \mathcal{A}$ approximating the expert policy $\pi_{\textrm{expert}}$. 
We define $\pi$ by a Convolutional Neural Network (CNN) parameterized by a set of weights $\theta$ and learned by minimizing the $L_2$ loss for the velocity controls $(\mathbf{v}_t, \bm{\omega}_t)$ and the cross-entropy loss $L_{\textrm{CE}}$ for the binary grasping signal $g_t$.
Given the state-action pair $(s_t, a_t)$ 
and the network prediction $\pi_\theta(s_t) = (\mathbf{\hat{v}}_t, \bm{\hat{\omega}}_t, \hat{g_t})$, we minimize the loss:
\begin{equation}
    \label{eq:bc}
    \lambda L_2\left([\mathbf{\hat{v}_t}, \bm{\hat{\omega}}_t], [\mathbf{v}_t, \bm{\omega}_t]\right) + (1-\lambda) L_{\textrm{CE}}\left(\hat{g_t}, g_t\right),
\end{equation}
where $\lambda \in [0, 1]$ is a scaling factor of the cross-entropy loss which we experimentally set to 0.9.

\subsection{Sim2Real transfer}
\label{sec:sim2real_eval}

Given a stochastic augmentation function $f$, we train a CNN $h$ to predict the cube position on a simulation dataset $\mathcal{D}^{\textrm{sim}}=\big\{(I^{\textrm{sim}}_i, p^{\textrm{sim}}_i)\big\}$.
Given an image $I^{\textrm{sim}}_i$, the function $f$ sequentially applies $N$ primitive transformations, each with a certain probability. This allows for bigger variability during the training.
We minimize the $L_2$ loss between the cube position $p^{\textrm{sim}}_i$ and the network prediction given an augmented depth image $h\left(f(I^{\textrm{sim}}_i)\right)$:
\begin{equation}
    \label{eq:train}
    \sum_k{\mathop{{}\mathbb{E}} L_2\left(h\left(f(I^{\textrm{sim}}_k)\right), p^{\textrm{sim}}_k\right)}.
\end{equation}
We evaluate augmentation functions by computing the average error of network prediction on a pre-recorded real-world dataset, $\mathcal{D}^{\textrm{real}}=\big\{(I^{\textrm{real}}_i, p_i^{\textrm{real}})\big\}$ as
\begin{equation}
    \label{eq:error}
    e^{real} = \frac{1}{n} \sum_{k=1}^n{L_2\left(h(I^{\textrm{real}}_k), p^{\textrm{real}}_k\right)}.
\end{equation}
The optimal augmentation function $f^{*}$ should result in a network $h$ with the smallest real-world error $e^{real}$. We assume that the same augmentation function will produce optimal control policies. We re-apply the learned stochastic function $f^{*}$ on individual frames of $\mathcal{D}^{\textrm{expert}}$ at every training epoch and learn $\pi^{\textrm{sim2real}}$. We use $\pi^{\textrm{sim2real}}$ to control the real robot using the same control actions as in the simulation, i.e., $a^{\textrm{real}}_t = (\mathbf{v}^{\textrm{real}}_t, \bm{\omega}^{\textrm{real}}_t, g^{\textrm{real}}_t) = \pi^{\textrm{sim2real}}(I^{\textrm{real}}_t)$, see Fig.~\ref{fig:overview}.

\subsection{Augmentation space}
\label{sec:augment_search_space}
We discretize the search space of augmentation functions by considering sequences of $N$ transformations from a predefined set. We then apply the selected sequence of transformations in a given order to each image. 
The predefined set of transformations consists of the depth-applicable standard transformations from PIL, a popular Python Image Library~\cite{PythonImagingLibrary},
as well as Cutout~\cite{devries2017cutout}, white (uniform) noise and salt (bernoulli) noise. 
We also take advantage of segmentation masks provided by the simulator and define two object-aware transformations, i.e., boundary noise and object erasing (see Section~\ref{sec:res_transforms_eval} for details). 
The identity (void) transformation is included in the set to enable the possibility of reducing $N$.
The full set of our eleven transformations is listed in Table~\ref{tab:transformations}.
Each transformation is associated with a magnitude and a probability of its activation. 
The magnitude defines the transformation-specific parameter.
For each transformation we define two possible magnitudes and three probabilities.
With $N=8$ in our experiments, our search space roughly includes $(11 \times 2 \times 3)^8 \approx 3.6 * 10^{14}$ augmentation functions.
We reduce the search space by restricting each transformation, except identity, to occur only once in any augmentation sequence.

\subsection{Real robot control}
\label{sec:robot_control}
\input{algos/mcts.tex}

To find an optimal sequence of augmentations, we use Monte Carlo Tree Search (MCTS)~\cite{Coulom2006EfficientSA} which is a heuristic tree search algorithm with a trade-off between exploration and exploitation. 
Our search procedure is defined in Algorithm~\ref{alg:autoaug}. The algorithm iteratively explores the Monte Carlo tree (lines 3-8) by sampling sequences of transformations (line 4), training a cube position prediction network on an augmented simulation dataset (line 5),  evaluating the trained network on the real dataset (line 6) and backpropagating the evaluation error through the Monte Carlo tree (line 7). Once the smallest error on the real dataset stays constant for 500 iterations, we choose the best augmentation function according to MCTS (line 9) and train sim2real control policies using the simulation dataset of augmented expert trajectories on the tasks of interest (line 10). Once trained, the control policies can be directly applied in real robot scenes without finetuning. The sequence of eight transformations found by MCTS is illustrated in Fig.~\ref{fig:transforms}.

%% file: figures/transforms.tex
\begin{figure*}[t]
\centering
 \includegraphics[width=\linewidth]{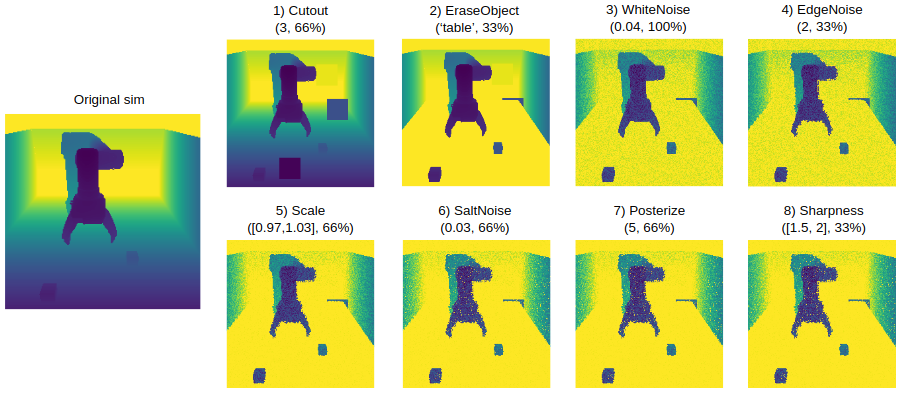}
\caption{The original synthetic depth image on the left is augmented by the sequence of eight random transformations learned by our method.}
\label{fig:transforms}
\end{figure*}

%% file: algos/mcts.tex
\begin{algorithm}
\caption{Sim2Real policy transfer algorithm}
\label{alg:autoaug}
\begin{algorithmic}[1]
\State *** Given datasets $\mathcal{D}^{\textrm{sim}}, \mathcal{D}^{\textrm{real}}, \mathcal{D}^{\textrm{expert}}_{\textrm{sim}}$ ***
\State $\textrm{MCTS} = \textrm{init\_mcts}()$
\Repeat
    \State $f = \textrm{MCTS.sample\_path}()$
    \State $\textrm{CNN} = \textrm{train\_cube\_prediction}(f, \mathcal{D}^{\textrm{sim}})$ \Comment{Eq.\ref{eq:train}}
    \State $e^{real} = \textrm{compute\_error}(\textrm{CNN}, \mathcal{D}^{\textrm{real}})$ \Comment{Eq.\ref{eq:error}}
    \State $\textrm{MCTS.update}(e^{real})$ \Comment{Backpropagate the error}
\Until{the smallest $e^{real}$ is constant for 500 iterations}
\State $f^{*} = \textrm{MCTS.select\_best\_path}()$
\State $\pi^{\textrm{sim2real}} = \textrm{train\_BC\_policy}(f^{*}, \mathcal{D}_{\textrm{sim}}^{\textrm{expert}})$ \Comment{Eq.\ref{eq:bc}}\\
\Return $\pi^{\textrm{sim2real}}$
\end{algorithmic}
\end{algorithm}

%% file: sections/results.tex
\section{RESULTS}

This section evaluates the transfer of robot control policies from simulation to real. First, we describe our tasks and the experimental setup in Section~\ref{sec:res_exp_setup}. We evaluate independently each of predefined basic transformations on the cube position prediction task in Section~\ref{sec:res_transforms_eval}. In Section~\ref{sec:res_augment_search}, we compare our approach of learning augmentation functions with baselines. Finally, we demonstrate the policy transfer to the real-world robotics tasks in Section~\ref{sec:res_real_robot}.

\subsection{Experimental setup}
\label{sec:res_exp_setup}

Our goal is to learn a policy to control a UR5 6-DoF robotic arm with a 3 finger Robotiq gripper for solving manipulation tasks in the real world. The policy takes as input 3 depth images $o_t \in \mathbb{R}^{H \times W \times 3}$ from the Kinect-1 camera positioned in front of the arm. We scale the values of depth images to the range $[0, 1]$. The policy controls the robot with an action $a_t \in \mathbb{R}^7$. The control is performed at a frequency of 10 Hz. All the objects and the robotic gripper end-effector are initially allocated within the area of $60 \times 60\, \text{cm}^2$ in front of the arm. The simulation environment is built with the \texttt{pybullet} physics simulator~\cite{Courmans2016} and imitates the real world setup. We consider three manipulation tasks.
\textbf{Cube picking task:} The goal of the task is to pick up a cube of size 4.7 cm and to lift it. In simulation, the cube size is randomized between 3 and 9 cm.
\textbf{Cubes stacking task:} The goal of the task is to stack a cube of size 3.5 cm on top of a cube of size 4.7 cm. We randomize the sizes of cubes in simulation between 3 to 9 cm.
\textbf{Cup placing task:} The goal of the task is to pick up a cup and to place it on a plate. In simulation, we randomly sample 43 plates from ModelNet~\cite{modelnet2015} and 134 cups from ShapeNet~\cite{shapenet2015}. We use three cups and three plates of different shapes in our real robot experiments.

\input{tables/transformations.tex}
\input{figures/tasks.tex}
\input{tables/regression.tex}
\input{tables/control_tasks_aug.tex}

\subsection{Evaluation of individual transformations}
\label{sec:res_transforms_eval}

Before learning complex augmentation functions, we independently evaluate each transformation from the set of transformations defined in Section~\ref{sec:augment_search_space}. In this section, we describe each transformation and the associated values of magnitude. \textbf{Affine} randomly translates and rotates the image in the range of $[-9,9]$ pixels and $[-5, 5]\degree$, or alternatively by 
$[-16,16]$ pixels and $[-10, 10]\degree$.
\textbf{Cutout}~\cite{devries2017cutout} samples one or three random rectangles in the image and sets their values to a random constant in $[0, 1]$. \textbf{Invert} function inverts each pixel value by applying the operation $x \mapsto 1-x$ and does not have any parameters. \textbf{Posterize} reduces the number of bits for each pixel value to be either 5 or 7. \textbf{Scale} randomly multiplies the image with a constant in one of the two ranges: $[0.95, 1.05]$ or $[0.97, 1.03]$. \textbf{Sharpness} increases the image sharpness either randomly in the range between $50\%$ and $100\%$ or by $100\%$. \textbf{WhiteNoise} adds uniform noise to each pixel with a magnitude of 0.04 or 0.08. \textbf{SaltNoise} sets each pixel value to 1 with the probability 0.01 or 0.03. \textbf{BoundaryNoise} uses the semantics mask of the simulator and removes patches of pixels located at the boundary between different objects. For BoundaryNoise, we remove either 2 or 4 pixels along the boundaries. \textbf{EraseObject} removes either the table or the walls behind the robot using the semantics segmentation mask. 
All the above transformations are associated with a probability in the set $\{33\%, 66\%, 100\%\}$. For instance, the probability $33\%$ means that the transformation is applied 1 out of 3 times and 2 our of 3 times it acts as the identity function.

As explained in Section~\ref{sec:sim2real_eval}, we evaluate each augmentation function by computing the prediction error of the cube position in real depth images after training the position regressor on simulated data. We collect 2000 pairs of simulated depth images and cube positions for training and 200 real depth images for evaluation. 
To be robust to viewpoint changes in real scenes, each simulated scene is recorded from five random viewpoints. We randomize the camera viewpoint in simulation around the frontal viewpoint by sampling the camera yaw angle in $[-15, 15]\degree$, pitch angle in $[15, 30]\degree$, and distance to the robot base in $[1.35, 1.50]$ m.
We only require the viewpoint of the real dataset to be within the simulated viewpoints distribution. Moreover, we allow to move the camera between different real robot experiments.
We treat each transformation in the given set as a separate augmentation function (sequence of length 1). We use each of them independently to augment the simulation dataset. 
Next, we train a CNN based on ResNet-18 architecture~\cite{He2016DeepRecognition} to predict the cube position given a depth image. During the network training, we compute the prediction error (\ref{eq:error}) on two validation datasets, 200 simulated images and 200 real images.
To evaluate each augmentation function more robustly, we always start the CNN training at the same initial position.
For each transformation, we report the cube position prediction error in Table~\ref{tab:transformations}. With no data augmentation, the trained network performs well in simulation (error of 0.63 cm) but works poorly on real images (error of 6.52 cm). Cutout transformation reduces the regression error by more than 4 cm and indicates that it should be potentially combined with other transformations.

\subsection{Augmentation function learning}
\label{sec:res_augment_search}

We compare the augmentation function learned by our approach to several baselines on the task of estimating the cube position in Table~\ref{tab:regression}. The baselines include training the network on synthetic images (i)~without any data augmentation, (ii)~with augmentation sequences composed of 8 random transformations (average over 10 random sequences), (iii) with a handcrafted augmentation sequence of four transformations built according to our initial intuition: Scale, WhiteNoise, EraseObjects and SaltNoise, (iv)~with the best single transformation from Section~\ref{sec:res_transforms_eval} and (v)~with the learned augmentation composed of 4 transformations.
To have a robust score and exclude outliers, we compute the median error over evaluations of 10 training epochs.

Baselines (i)-(iii) with no augmentation learning demonstrate worst results on the real dataset.
Results for learned transformations (iv)-(vi) show that more transformations with different probabilities help to improve the domain transfer. Table~\ref{tab:regression} also demonstrates the trade-off between the performance in different domains: the better augmentation works on the real dataset, the worse it performs in the simulation.
Effectively, the learned augmentation shifts the distribution of simulated images towards the distribution of real images. As a consequence, the network performs well on the real images that are close to the training set distribution and works worse on the original simulated images that lie outside of the training set distribution.
The best augmentation sequence found by our method is illustrated in Fig.~\ref{fig:transforms} and contains the following transformations: Cutout, EraseObject, WhiteNoise, EdgeNoise, Scale, SaltNoise, Posterize, Sharpness. Learning an augmentation function of length 8 takes approximately 12 hours on 16 GPUs. The vast majority of this time is used to train the position estimation network while MCTS path sampling, evaluation and MCTS backpropagation are computationally cheap. We iteratively repeat the training and evaluation routine until the error does not decrease for a sufficiently long time (500 iterations). Once the sim2real augmentation is found, it takes approximately an hour to train the BC control policy.

\subsection{Real robot control}
\label{sec:res_real_robot}

In this section we demonstrate that the data augmentation learned for a proxy task transfers to other robotic control tasks. We collect expert demonstrations where the full state of the system is known and an expert script can easily be generated at training time. We augment the simulated demonstrations with the learned data augmentation and train BC policies without any real images.  Moreover, we show that our augmentation is not object specific and transfers to tasks with new object instances not present in the set of expert demonstrations. For each task, we compare the learned augmentation function with 3 baselines: no augmentation, handcrafted augmentation and best single transformation. Each evaluation consists of 20 trials with random initial configurations. The results are reported in Table~\ref{tab:control_pick_aug}.

\textbf{Cube picking task.} Success rates for policies learned with different augmentation functions are strongly correlated with results for cube position estimation in Table~\ref{tab:regression}. The policy without augmentation has a success rate 3/20. Single transformation and handcrafted augmentation have 9/20 and 8/20 successful trials respectively. The sim2real policy learned with our method succeeds 19 out of 20 times.
 
\textbf{Cube stacking task.} Given a more difficult task where more precision is required, the baseline approaches perform poorly and achieve the success rate of only 2/20 for the handcrafted augmentation. We observe most of the failure cases due to imprecise grasping and stacking. We successfully tested the learned data augmentation function on cubes of varying sizes which indicates high control precision. Overall, our method was able to stack cubes in 18 runs out of 20.

\textbf{Cup placing.} Solving the Cup placing task requires both precision and the generalization to previously unseen object instances. The policies are trained over a distribution of 3D meshes and thus leverage the large dataset available in the simulation. All baselines fail to solve the Cup placing except for the handcrafted augmentation which succeeds 6 times out of 20. Our approach is able to solve the task with the success rate of 15/20 despite the presence of three different instances of cups and plates never seen during training.
These results confirm our hypothesis that the augmentations learned for a proxy task of predicting the cube position, generalize to new objects and tasks.

%% file: tables/transformations.tex
\begin{table}[]
\centering
\begin{tabular}{lcc}
\toprule
Transformation & Error in sim & Error in real \\
\midrule
Identity & 0.63 $\pm$ 0.50 & 6.52 $\pm$ 5.04 \\
Affine & \textbf{0.59 $\bm{\pm}$ 0.45} & 4.83 $\pm$ 4.42 \\
Cutout & 1.19 $\pm$ 0.87 & \textbf{1.86 $\bm{\pm}$ 2.45} \\
Invert & 0.88 $\pm$ 0.60 & 4.63 $\pm$ 3.28 \\
Posterize & 0.66 $\pm$ 0.48 & 5.54 $\pm$ 4.59 \\
Scale & 0.67 $\pm$ 0.47 & 6.00 $\pm$ 4.37 \\
Sharpness & 0.83 $\pm$ 0.49 & 5.48 $\pm$ 3.84 \\
WhiteNoise & 0.68 $\pm$ 0.54 & 3.60 $\pm$ 2.33 \\
SaltNoise & 0.72 $\pm$ 0.50 & 2.42 $\pm$ 1.16 \\
BoundaryNoise & 0.88 $\pm$ 0.66 & 2.06 $\pm$ 1.17 \\
EraseObject & 0.64 $\pm$ 0.47 & 1.93 $\pm$ 1.01 \\
\bottomrule
\end{tabular}
\caption{
Cube prediction error (in cm) for synthetic and real depth images evaluated separately for each of eleven transformations considered in this paper. The errors are averaged over 200 pairs of images and cube positions.}
\label{tab:transformations}
\end{table}

%% file: figures/tasks.tex
\begin{figure*}[t]
    \centering
    \begin{subfigure}[b]{0.475\textwidth}
        \centering
        \includegraphics[width=\textwidth]{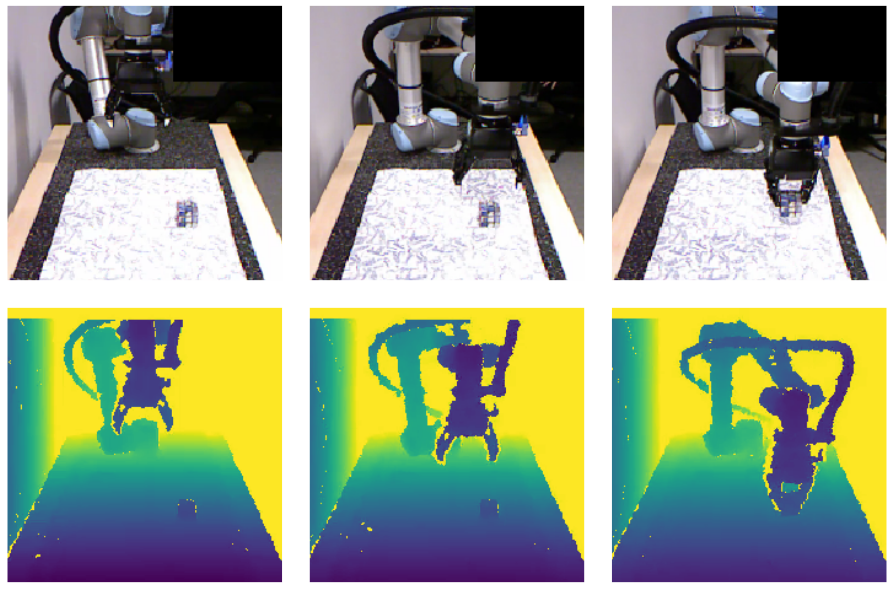}
        \caption[]%
        {{\small Cube picking}}    
        \label{fig:cube_pick}
    \end{subfigure}
    \quad
    \begin{subfigure}[b]{0.475\textwidth}  
        \centering 
        \includegraphics[width=\textwidth]{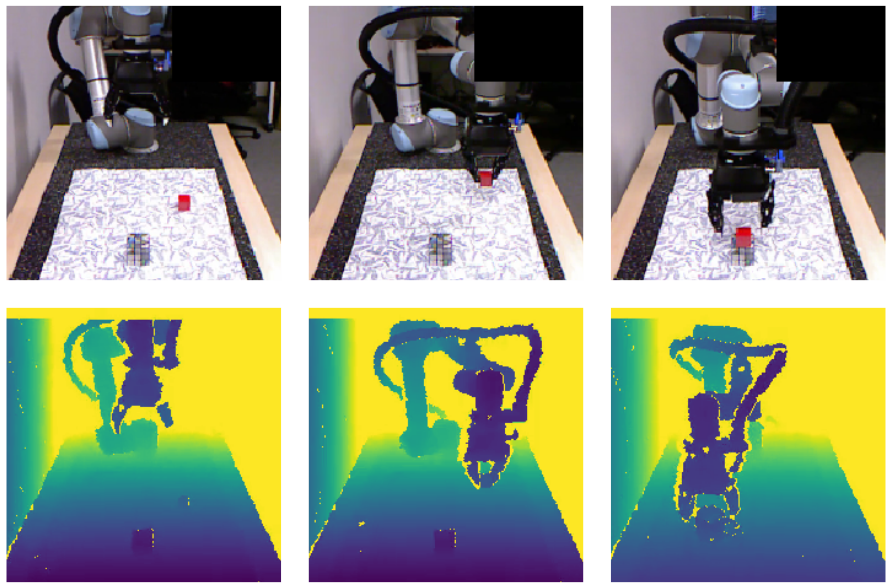}
        \caption[]%
        {{\small Cube stacking}}    
        \label{fig:cube_stack}
    \end{subfigure}
    \vskip\baselineskip
    \begin{subfigure}[b]{0.475\textwidth}   
        \centering 
        \includegraphics[width=\textwidth]{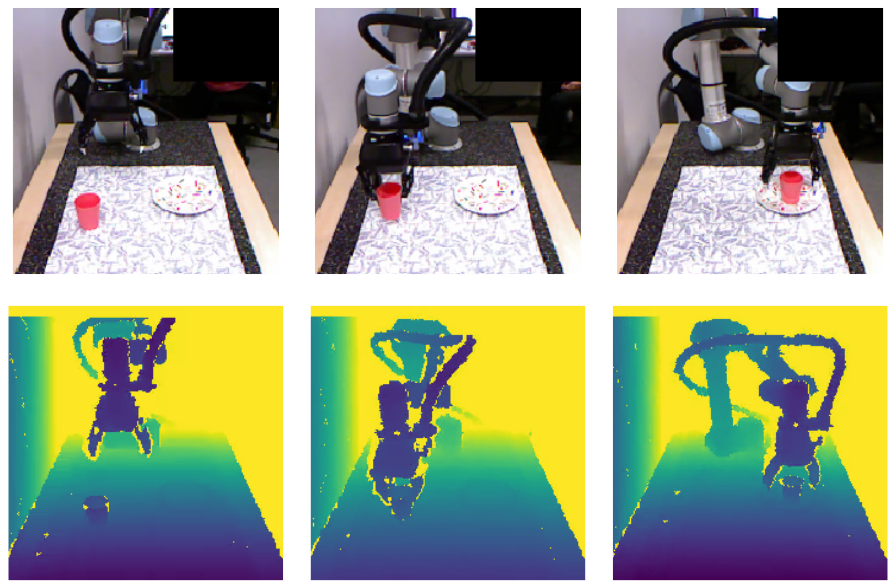}
        \caption[]%
        {{\small Cup placing}}    
        \label{fig:cup_place1}
    \end{subfigure}
    \quad
    \begin{subfigure}[b]{0.475\textwidth}   
        \centering 
        \includegraphics[width=\textwidth]{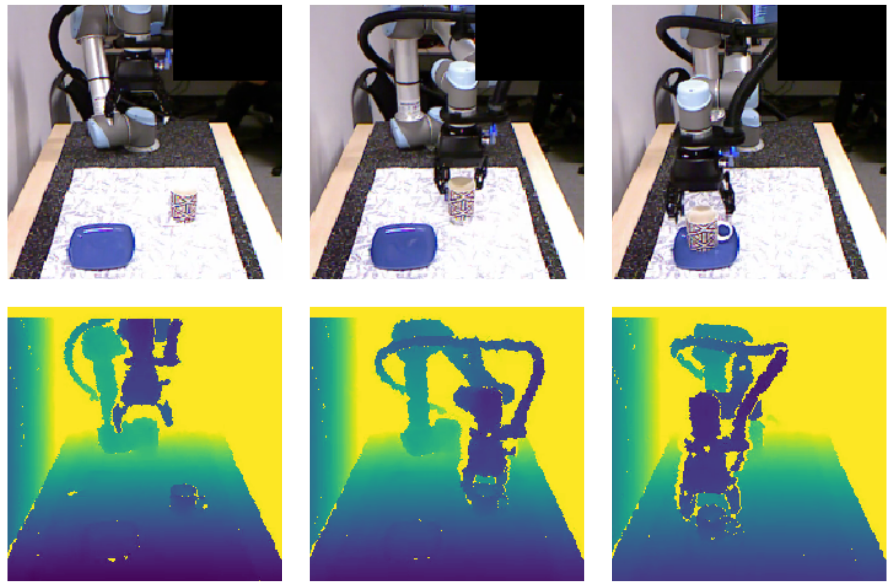}
        \caption[]%
        {{\small Cup placing}}    
        \label{fig:cup_place2}
    \end{subfigure}
    \caption{Frame sequences of real world tasks performed with a policy learned on a simulation dataset augmented with our approach. The tasks are: a) picking up a cube, b) stacking two cubes, c) placing a cup on top of a plate, d) placing another cup on top of another plate.}
    \label{fig:tasks}
\end{figure*}

%% file: tables/regression.tex
\begin{table}[]
\setlength\tabcolsep{3pt} 
\centering
\begin{tabular}{llcc}
\toprule
&Augmentation &  Error in sim & Error in real \\
\midrule
i&None & \textbf{0.63 $\bm{\pm}$ 0.50} & 6.52 $\pm$ 5.04 \\
ii&Random (8 operations) & 6.56 $\pm$ 4.05 & 5.77 $\pm$ 3.12 \\
iii&Handcrafted (4 operations) & 0.99 $\pm$ 0.68 & 2.35 $\pm$ 1.36 \\
iv&Learned (1 operation) & 1.19 $\pm$ 0.87 & 1.86 $\pm$ 2.45 \\
v&Learned (4 operations) & 1.21 $\pm$ 0.78 & 1.17 $\pm$ 0.71 \\
vi&Learned (8 operations) & 1.31 $\pm$ 0.90 & \textbf{1.09 $\bm{\pm}$ 0.73} \\
\bottomrule
\end{tabular}
\caption{Cube prediction error (in cm) on synthetic and real depth images using different types of depth data augmentation. 
More augmentations increase the error for synthetic images and decrease the error for real images as expected from our optimization procedure. The errors are averaged over 200 pairs of images and cube positions.}
\label{tab:regression}
\end{table}

%% file: tables/control_tasks_aug.tex
\begin{table}[]
\centering
\begin{tabular}{lcccc}
\toprule
Augmentation  &  Pick & Stack & Cup Placing\\
\midrule
None &  3/20 & 1/20 & 0/20\\
Handcrafted (4 operations) & 9/20 & 2/20 & 6/20\\
Learned (1 operation) & 8/20 & 1/20 & 1/20\\
Learned (8 operations) & \textbf{19/20} & \textbf{18/20} & \textbf{15/20}\\
\bottomrule
\end{tabular}
\caption{Success rates for control policies executed on a real robot (20 trials per experiment). Results are shown for three tasks and alternative depth image augmentations.
} 
\label{tab:control_pick_aug}
\end{table}

%% file: sections/conclusion.tex
\section{CONCLUSIONS}

In this work, we introduce a method to learn augmentation functions for sim2real policy transfer. To evaluate the transfer, we propose a proxy task of object position estimation that requires only a small amount of real world data. Our evaluation of data augmentation shows significant improvement over the baselines. We also show that the performance on the proxy task strongly correlates with the final policy success rate. Our method does not require any real images for policy learning and can be applied to various manipulation tasks. We apply our approach to solve three real world tasks including the task of manipulating previously unseen objects.\medskip \\

\noindent
{\em Acknowledgements.} This work was supported in part by the ERC grants ACTIVIA and ALLEGRO, the Louis Vuitton / ENS Chair on Artificial Intelligence and the DGA project DRAAF.